\documentclass{article}

\usepackage[preprint]{neurips_2024}

\usepackage[utf8]{inputenc} 
\usepackage[T1]{fontenc}    
\usepackage{hyperref}       
\usepackage{url}            
\usepackage{booktabs}       
\usepackage{amsfonts}       
\usepackage{nicefrac}       
\usepackage{microtype}      
\usepackage{amsmath,amssymb} 
\usepackage{color}
\usepackage{hyperref}
\usepackage{bm}
\usepackage{multicol,multirow}
\usepackage{threeparttable}
\usepackage{algorithm,algorithmic}
\usepackage{graphicx}
\usepackage{arydshln}
\usepackage{booktabs}
\usepackage{listings}
\usepackage[table,xcdraw]{xcolor}
\usepackage{wrapfig}
\usepackage[misc,geometry]{ifsym}
\usepackage[marginal]{footmisc}
\usepackage{amsthm}
\usepackage{makecell}

\newcommand{\secref}[1]{Section~\ref{sec:#1}}
\newcommand{\figref}[1]{Figure~\ref{fig:#1}}
\newcommand{\tabref}[1]{Table~\ref{tab:#1}}
\newcommand{\equref}[1]{Eq.~\ref{equ:#1}}

\title{TTAQ: Towards Stable Post-training Quantization in \\ Continuous Domain Adaptation}

\author{%
  Junrui Xiao\textsuperscript{1,2}, Zhikai Li\textsuperscript{1,2}, Lianwei Yang\textsuperscript{1,2}, Yiduo Mei\textsuperscript{3}, Qingyi Gu\textsuperscript{1$*$} \\
  \textsuperscript{1}Institute of Automation, Chinese Academy of Sciences. Beijing, China \\
  \textsuperscript{2}School of Artificial Intelligence, University of Chinese Academy of Sciences. Beijing, China \\
  \textsuperscript{3}Inspur Yunzhou Industrial Internet Co., Ltd. Jinan City, Shandong Province, China\\
  \texttt{\{xiaojunrui2020, qingyi.gu\}@ia.ac.cn}
}

\begin{document}

\maketitle

\begin{abstract}
Post-training quantization (PTQ) reduces excessive hardware cost by quantizing full-precision models into lower bit representations on a tiny calibration set, without retraining.
Despite the remarkable progress made through recent efforts, traditional PTQ methods typically encounter failure in dynamic and ever-changing real-world scenarios, involving unpredictable data streams and continual domain shifts, which poses greater challenges.
In this paper, we propose a novel and stable quantization process for test-time adaptation (TTA), dubbed TTAQ, to address the performance degradation of traditional PTQ in dynamically evolving test domains.
To tackle domain shifts in quantizer, TTAQ proposes the Perturbation Error Mitigation (PEM) and Perturbation Consistency Reconstruction (PCR). Specifically, PEM analyzes the error propagation and devises a weight regularization scheme to mitigate the impact of input perturbations. On the other hand, PCR introduces consistency learning to ensure that quantized models provide stable predictions for same sample. 
Furthermore, we introduce Adaptive Balanced Loss (ABL) to adjust the logits by taking advantage of the frequency and complexity of the class, which can effectively address the class imbalance caused by unpredictable data streams during optimization.
Extensive experiments are conducted on multiple datasets with generic TTA methods, proving that TTAQ can outperform existing baselines and encouragingly improve the accuracy of low bit PTQ models in continually changing test domains.  For instance, TTAQ decreases the mean error of 2-bit models on ImageNet-C dataset by an impressive 10.1\%.
\end{abstract}

\section{Introduction}
In recent years, deep neural networks (DNNs) have achieved remarkable progress in various computer vision tasks, such as image classification~\cite{CrossViT,ViT,LanchantinWOQ21}, object detection~\cite{CarionMSUKZ20,PAOD,Xiao2023DCIFPNDC,back2}, and semantic segmentation~\cite{StrudelPLS21,ZhengLZZLWFFXT021,back1}, thanks to the success of large-scale models trained on extremely large datasets.
However, increasing demands for computation, memory, and storage present challenges in deploying these models in resource-limited devices or environments. Consequently, various model compression techniques have been proposed to address these resource constraints and accelerate inference, including pruning~\cite{he2018soft,HWCCC22}, knowledge distillation~\cite{LinXWYCLW22,back3,romero2014fitnets,xu2022ida}, structured factorization and quantization~\cite{Q-ViT,repqvit,LiuWHZMG21,Xiao2023PatchwiseMQ,Quant1}.

\begin{figure}[t]
  \centering
  \vspace{0.15in}
  \includegraphics[width=0.8\linewidth]{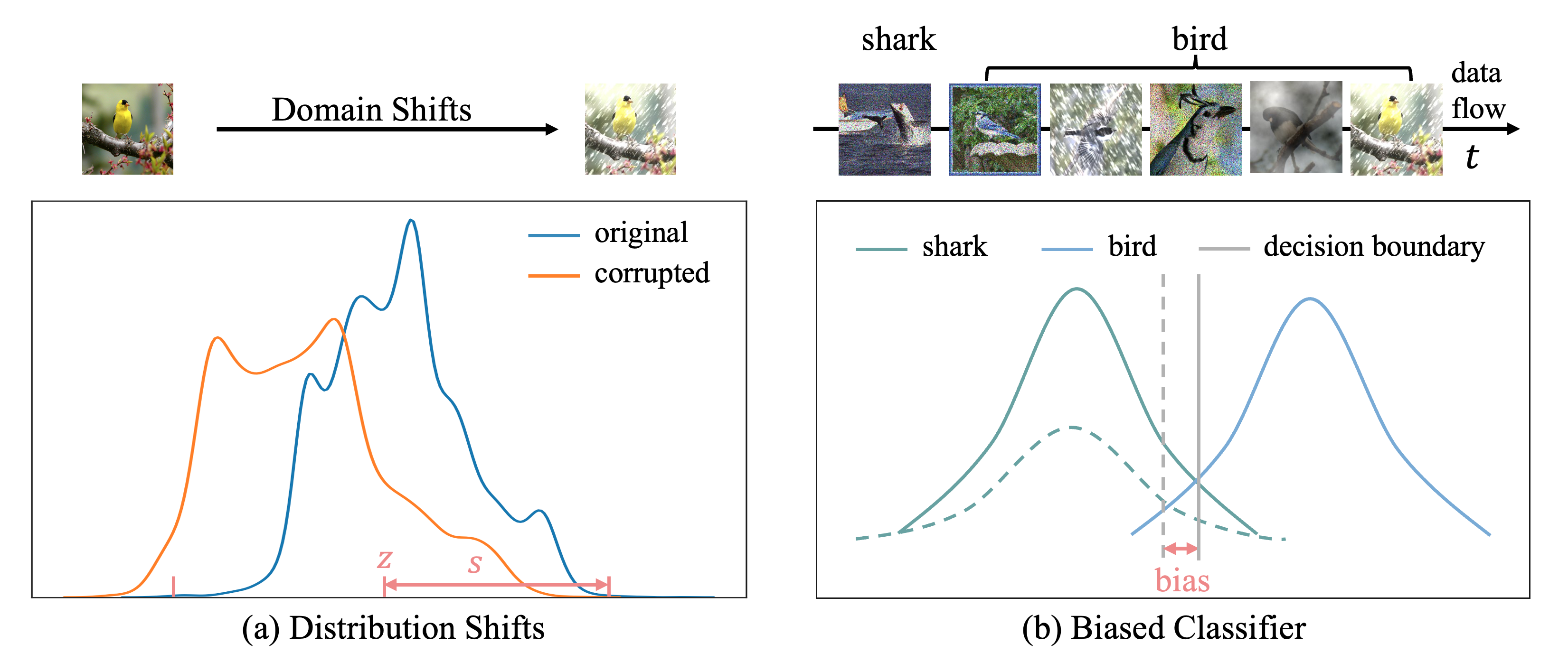}
  \caption{Illustration of the problem: (a) The distribution of the 1st to 10th channels of the activation in ResNet-18's first convolutional layer shows a significant shift. The quantization parameters calibrated based on the original distribution are inaccurate, resulting in performance degradation; (b) A toy example illustrating the bias introduced by class-imbalanced samples. The solid gray line represents the unbiased classifier with the optimal decision boundary, while the dotted line depicts the learned biased classifier on imbalanced streaming samples, which also hinders the accuracy.}
  \vspace{-0.15in}
  \label{fig:drawback}
\end{figure}
Model quantization has garnered significant attention as an effective technique that converts the weights and activations of large models from full-precision to low-bit fixed-point representations.  Most quantization works can be divided into two categories: quantization-aware training (QAT)~\cite{DBLP:conf/icml/NagelFBB22, choi2018pact, esser2019learned} and post-training quantization (PTQ)~\cite{repqvit, FQ-ViT, LiuWHZMG21, Yuan2022PTQ4ViTPQ, PTQforViT}. By retraining the entire labeled dataset, QAT typically possesses higher accuracy but requires significant memory consumption and GPU efforts. On the contrary, PTQ only requires a tiny unlabeled calibration set to quantize the pre-trained parameters without retraining, and thus is regarded as a practical solution in widespread scenarios. Existing PTQ methods demonstrate great potential in terms of both accuracy and efficiency, as they can achieve good prediction accuracy on domain-invariant~\cite{wang2022exploring} test sets even when the quantized weights are reduced to 4 bits.

However, in real-world applications, domain shifts frequently occur at test-time due to corruptions, changes in weather conditions, camera sensor differences, and other factors. When there is an inconsistency between the distributions of the test and training data, the conventional PTQ method, which is calibrated on a fixed distribution, experiences severe performance degradation. Moreover, the data collected from edge devices, such as Internet of Things (IoT) products or wearable devices, arrives in a sequential manner, which presents additional challenges. This is due to the limited computational resources and the need for real-time processing, which makes it impractical to retrain all parameters using streaming data.

To perform online adaptation on the test stream characterized by an unpredictable distribution and provide instant predictions for incoming test samples, Continual Test-Time Adaptation (CTTA) ~\cite{wang2022continual, gong2022note, dobler2023robust, song2023ecotta} has emerged as a robust solution. CoTTA introduces an Exponentially Moving Average (EMA)~\cite{cai2021exponential} teacher-student architecture to self-train all parameters using pseudo labels, which necessitates substantial memory during adaptation. To enhance efficiency, EATA~\cite{Niu2022EfficientTM} employs an active sample selection criterion to identify reliable and non-redundant samples for updating the parameters of the batch normalization layer. However, current TTA approaches have focused primarily on full-precision models. The issue of \textit{how to achieve stable quantization in CTTA} has not yet been explored thoroughly.

In this work, we first empirically reveal the potential problems of PTQ in the CTTA setting from two perspectives. Firstly, we observe that domain shifts directly cause alterations in the activation distribution, resulting in inaccurate calibrated quantization parameters (e.g., scale and zero point) and an increase in quantization errors, as illustrated in \figref{drawback}(a). Secondly, when adapting on streaming data with an unpredictable distribution, imbalanced class distributions can easily arise, as shown in \figref{drawback}(b), leading to a decrease in the model's generalization capability and resulting in catastrophic forgetting of the minor class~\cite{McCloskey1989CatastrophicII}.

Motivated by the analysis above, we propose a stable PTQ scheme in continuous domain adaptation, namely TTAQ, which considers both the robustness of quantization parameters and the imbalance of classes in adaptation. 
First, we conduct an analysis of the error propagation stemming from perturbations, considering the signal-to-noise ratio, and then propose the Perturbation Error Mitigation (PEM), which introduces a weight-regularization scheme to mitigate the effect of input perturbations. 
Second, to further enhance the robustness, we propose Perturbation Consistent Reconstruction (PCR), which adds small perturbations during the block-wise reconstruction process and encourages quantized model can have the same output through perturbation consistency loss.
Finally, we introduce Adaptive Balanced Loss (ABL) to address the class imbalance issue caused by unpredictable data streams. ABL reweights the samples based on class-wise frequency and adjusts logits by the accumulated gradients of each class, which effectively promotes the learning of unbiased classifiers.

We summarize our main contributions as follows:
\begin{itemize}
\item To the best of our knowledge, the proposed TTAQ is the first attempt to achieve stable post-training quantization in continuous domain adaptation by considering both quantization robustness and class imbalance.
\item To enhance robust quantization, we introduce PEM and PCR, which effectively mitigate the impact of activation perturbations and improve prediction consistency.
\item For addressing class imbalance, we propose ABL, which reweights samples based on class-wise factors that incorporate the frequency and complexity of each class.
\item We conduct evaluations of TTAQ on various vision tasks, including image classification, object detection, and instance segmentation. Encouragingly, TTAQ demonstrates superior performance compared to existing PTQ approaches.
\end{itemize}

\section{Related works}
\subsection{Model Quantization.}
Model quantization~\cite{gholami2022survey, nagel2021white, krishnamoorthi2018quantizing} is one of the effective techniques for compressing and deploying neural networks, which quantizes the full-precision parameters to lower bit-width, and can be categorized as Quantization-Aware Training (QAT) and Post-Training Quantization (PTQ). QAT~\cite{DBLP:conf/icml/NagelFBB22, choi2018pact, esser2019learned} shows the potential of low-precision computation by retraining the quantized model on a labeled training dataset. However, the time-consuming retraining and complex hyperparameters lead to limited applications. PTQ~\cite{repqvit, FQ-ViT, LiuWHZMG21, Yuan2022PTQ4ViTPQ, PTQforViT} stands out as an efficient and practical compression approach, as it directly quantizes the model without retraining. Adaround\cite{nagel2020up} introduces a continuous variable to each weight value to determine whether it should be rounded up or down, instead of using the nearest rounding.
Brecq\cite{li2021brecq} first reconstructs the quantization parameters block by block and determines the step size for activation after the reconstruction stage. QDrop~\cite{wei2022qdrop} incorporates the activation error into the reconstruction process and introduces a drop operation to improve quantization. Pd-Quant~\cite{liu2023pd} argues that considering only local information leads to sub-optimal quantization parameters. Therefore, it proposes using the information of differences between network predictions before and after quantization to determine the quantization parameters. 

Although existing PTQ models, which were calibrated on a fixed distribution of data, have achieved significant performance, it remains challenging to ensure their reliable prediction in dynamic environments. In particular, these environments are characterized by ever-changing streaming data.
\subsection{Continual Test-time Adaptation.}
In recent times, there has been a remarkable increase in research interest pertaining to test-time domain adaptation~\cite{wang2022continual, varsavsky2020test, gao2022visual} that adapts models during inference for addressing the domain shift problem. 
Test-time adaptation involves considering a fixed scenario where the test data are provided in batch format. NORM~\cite{Mirza2021TheNM} adjusts the statistics of batch normalization (BN) layers using the current test samples, without updating other parameters. TENT~\cite{wang2020tent} introduces entropy minimization as a test-time optimization objective. SAR~\cite{niu2023towards} proposes a reliable and sharpness-aware entropy minimization method to remain robust in the presence of large and noisy gradients. 
Continual test-time adaptation is a scenario where the target domain is dynamic with a stream of continually changing samples, presenting increased challenges for traditional computer vision tasks. 
CoTTA~\cite{wang2022continual} employs weighted averaging, improved average prediction, and random parameter recovery to prevent catastrophic forgetting. EcoTTA~\cite{song2023ecotta} introduces a small meta-network to regularize the outputs from both the meta-network and the frozen network. RMT~\cite{dobler2023robust} introduces a symmetric cross-entropy loss through gradient analysis. RoID\cite{marsden2024universal} defines all relevant settings as universal TTA and then proposes a model-agnostic certainty and diversity weighting.

In this paper, our aim is to achieve a stable and efficient PTQ under a continual every-changing domain, without the need for additional prompts or adapters.

\begin{figure*}[t]
  \centering
  \vspace{0.14in}
  \includegraphics[width=1\linewidth]{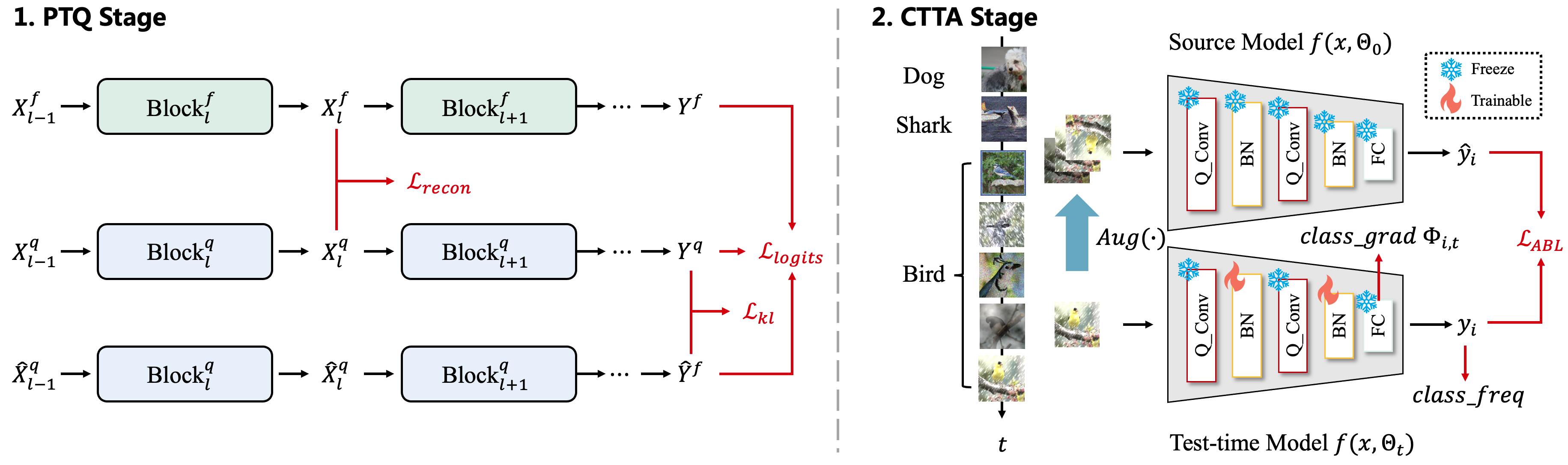}
  \vspace{-0.25in}
  \caption{Pipeline of TTAQ. In Post-Traing Quantization stage (left), perturbation consistency reconstruction apply consistency learning by encouraging quantized model can have the same output when there is a slight perturbation between $X^q_{l-1}$ and $\hat{X}^q_{l-1}$. TTAQ also introduce global information by $\mathcal{L}_{logits}$ to alleviate the overfitting problem.
  In Continual Test-Time Adaption stage (right), adaptive balanced loss adjusts the logits by the frequency and accumulated gradient of each class to enhance the learning of unbiased classifiers.}
  \label{fig:overview}
\end{figure*}
\section{Method}
In this section, we introduce the proposed TTAQ in detail. To reduce the perturbation from domain shifts, we propose perturbation error mitigation and perturbation consistency reconstruction, respectively, as detailed in \secref{PEM} and \secref{PCR}. To address the class imbalance of unpredictable streaming data, we propose adaptive balanced loss in \secref{ABL}. The overview of TTAQ is shown in \figref{overview}.
\subsection{Preliminaries}\label{sec:Pre}
\textbf{Overview of CTTA.}
In this work, our focus lies on the continuous adaptation environment during testing time, where access to the source data is not available during adaptation. Here, the pre-trained model must adapt to an unlabeled and continuously changing target domain in an online manner. 
First, consider a model $f\left(x_i ; \Theta\right) \in \mathbb{R}^{H \times W \times C}$ trained using data $D_{\text {source}}=\left\{\left(x_i, y_i\right)\right\}_{i=1}^N$, where $\left(x_i, y_i\right)$ denote the image and label $i_{th}$ drawn from the source distribution $D_{\text {source}}$, respectively. Now, deploy the pre-trained model $f\left(x_i ; \Theta\right)$ to the test environments where the test data at inference time T is $D^T_{\text {target}}=\left\{x_j\right\}_{j=1}^B$, and $D^T_{\text {target}}$ deviates from the i.i.d. assumption.
Our goal is to adapt the model $f\left(\cdot; \Theta_t\right)$ to $f\left(\cdot; \Theta_{t+1}\right)$ using only the unlabeled input $x_j^T$ while making predictions directly. Following previous CTTA methods~\cite{Niu2022EfficientTM,marsden2024universal}, we freezes the majority of the model’s parameters, permitting only minor adjustments for efficient adaptation.

\textbf{Scheme of PTQ.}
Post-training quantization (PTQ) leverages a subset of unlabeled images to determine the scaling factors $s$ and zero point $z$ of activations and weights for each layer. For convenience, we revisit the uniform quantization. Uniform quantization quantizes the weights or activation values from full precision $X$ to integer ${X^q}$ when given a quantization bit width $b$. The processes quantization and dequantization are shown as follows:
\begin{align}
\label{equ:quant}
  Quant&: \bm{X}^{q} = \text{clamp}\left(\left\lfloor \frac{\bm{X}}{s} \right\rceil+z, 0, 2^b-1 \right) \\
\label{equ:dequant}
  DeQuant&: \bm{X}^{dq} = s\left(\bm{x}^{(\mathbb{Z})}-z\right) \approx \bm{x}
\end{align}
where $\left\lfloor\cdot\right\rceil$ represents the round function, clamp limits the range of values to between 0 and $2^b-1$. $s$ and $z$ can be caculated as:
\begin{align}
\label{equ:sz}
  s = \frac{\max(\bm{X})-\min(\bm{X})}{2^b-1}, \quad z = \left\lfloor-\frac{\min(\bm{X})}{s} \right\rceil.
\end{align}

Block-wise\cite{nagel2020up, wei2022qdrop} reconstruction is proposed to further to minimize the quantization error of PTQ. In this work, we also employ the block-wise reconstruction as the learning objective. For the $l_{th}$ block of the full-precision model and quantized model, we achieve this by minimizing the error between the block outputs. The process can be formulated as: 
\begin{align}
\label{equ:reconstruction}
\arg {\underset{s_l}{\min}} \mathcal{L}_{recon}({X}_l^f, X_l^q)  
\end{align}
where the ${X}_l^f$ and $X_l^q$ represent the output of the $l_{th}$ block of the full-precision model and the quantized model, respectively.


\subsection{Perturbation Error Mitigation}\label{sec:PEM}
In this section, we detail perturbation error mitigation. Following EATA~\cite{Niu2022EfficientTM}, we adapt the model only by updating the parameters in the Batch Norm layer instead of the whole model to prioritize efficiency for edge devices. However, during adaptation to the changing target domain, the distribution of the input to the PTQ model will change (as shown in \figref{drawback}(a)), and the quantization parameters cannot be updated, resulting in the layer-by-layer accumulation of quantization errors. As revealed in \tabref{evidence}, we also conduct a experiment to show that the performance of the PTQ model degrades more under the CTTA setting.

To address the above issues, we first conduct an analysis of the error propagation resulting from perturbations\cite{Meller2019SameSB,Park2022SymmetryRA}. Consider a convolutional layer with kernel $W^{ij}_{\left(K_x \cdot K_y \cdot F_{\text {in }}\right) \times F_{\text {out }}}$, input $x^{j}_{1\times \left(K_x \cdot K_y \cdot F_{\text {in }}\right)}$, and output $y^{i}_{1\times F_{\text {out}}}$. For notional simplicity we model both convolution and linear operation as matrix multiplication and model the domain shift as $g(\cdot)$:
\begin{align}
\label{equ:1}
y_i=\sum_j^N W_{i, j} g(x_j)
\end{align}
where $W_{i, j}$ is an arbitrary and fixed value, and $g(x_j)$ represents a corrupted activation assumed to be an i.i.d. variable. When quantizer $Q(\cdot)$ is applied to the weight, the output can be formulated as:
\begin{align}
\label{equ:2}
\hat{y}_i=\sum_j^N Q(W_{i, j}) g(x_j)
\end{align}

We use signal noise ratio to compute the effect of the input perturbation:
\begin{align}
\label{equ:3}
S N R_{y_i} = \frac{E_i[(\hat{y}_i-y_i)^2]}{E_i[y_i^2]}
\end{align}
\begin{table}[t]
\centering
\caption{Mean Error (\%) of Quantized ResNet-18~\cite{He2015DeepRL} with different bit-widths on ImageNet-C~\cite{Hendrycks2019BenchmarkingNN}. CTTA Error represents the performance degradation of the PTQ Model in CTTA setting, while Quant Error represents the performance degradation of the PTQ Model in the source domain compared to FP model. As is evident, there is an extra degradation caused by inaccurately calibrated quantization parameters.}
\scalebox{0.82}{
\begin{tabular}{@{}c|ccccc@{}}
\toprule
Method & Bit & Mean & CTTA   Error & Quant   Error & Extra Error \\ \midrule
\multirow{3}{*}{TENT} & FP & 69.69 & - & - & - \\
 & W4A4 & 75.69 & -6 & -4.03 & \textbf{-2.03} \\
 & W2A4 & 80.86 & -12.83 & -10 & \textbf{-2.83} \\ \midrule
\multirow{3}{*}{EATA} & FP & 65.9 & - & - & - \\
 & W4A4 & 72.8 & -6.9 & -4.03 & \textbf{-1.87} \\
 & W2A4 & 78.12 & -12.22 & -10 & \textbf{-2.22} \\ \bottomrule
\end{tabular}
}
\label{tab:evidence}
\end{table}
where $y_i$ and $\hat{y}_i$ denote the $i_{th}$ output with the full-precision weight $W_{i, j}$ and quantized weight $Q(W_{i, j})$, respectively. Since the $E_i[y_i^2]$ term is constant for $i_{th}$ layer, the optimization objective is to minimize the $E_i[(\hat{y}_i-y_i)^2]$.
\begin{equation}
\label{equ:objective}
\begin{aligned}
&\arg {\underset{Q(\cdot)}{\min}} E_i[(\hat{y}_i-y_i)^2]=\arg {\underset{Q(\cdot)}{\min}}(E_i[(\hat{y}_i-y_i)])^2+Var(\hat{y}_i-y_i)\\
\end{aligned}
\end{equation}

Let $\mathbb{E}\left(g\left(x_i\right)\right)=\mu_g$ and $\operatorname{Var}\left(g\left(x_i\right)\right)=\sigma_g^2$, the expected value and the variance of output $y_i$ can be expressed as:
\begin{equation}
\mathbb{E}_i\left(y_i\right)=N \mu_g \mathbb{E}_i(W_{i,j}),\quad \operatorname{Var}_i\left(y_i\right)=N \sigma_g^2 \mathbb{E}_i(W^2_{i,j})
\end{equation}

It is straightforward to show that the optimization objective in \equref{objective} can be rewritted as:
\begin{equation}
\label{equ:final}
\arg {\underset{Q(\cdot)}{\min}}[\mathbb{E}_i(Q(W_{i,j})) - \mathbb{E}_i(W_{i,j})]^2+[\mathbb{E}_i(Q(W_{i,j})^2) - \mathbb{E}_i(W_{i,j}^2)]
\end{equation}

However, rigorously solving the final objective in \equref{final} becomes challenging since PTQ only conducts a simple parameter space search with a limited calibration set. To ease the difficulty, we adopt the additional condition as given by:
\begin{equation}
\label{equ:rewritte}
\mathbb{E}_i(Q(W_{i,j}))=\mathbb{E}_i(W_{i,j}),\quad \mathbb{E}_i(Q(W_{i,j})^2)=\mathbb{E}_i(W_{i,j}^2)
\end{equation}

To this end, we can mitigate perturbation error by regularizing the distribution of weights to satisfy \equref{rewritte}. Inspired by the crucial role of normalization in quantization\cite{li2019additive,Banner2018ACIQAC,Chen2023OvercomingFC}, we devise a weight regularization scheme:
\begin{equation}
\label{equ:reg}
Q(\hat{W}_{i, j})=Q(\alpha \cdot \frac{W_{i, j}-\mu_{W_{i, \cdot}}}{\sigma_{W_{i, .}}})
\end{equation}
where the mean $\mu_{W_{i, \cdot}}$ and variance $\sigma_{W_{i, .}}$ are computed by $\mu_{W_{i, .}}=\frac{1}{N} \sum_j^N W_{i, j}$ and $\sigma_{W_{i, .}}^2=\frac{1}{N} \sum_j^N W_{i, j}^2-\mu_{W_i}^2$, while $\alpha$ is a fixed constant. By recalling \equref{rewritte}, we can find that $\mathbb{E}_i(Q(\hat{W}_{i,j})=\mathbb{E}_i(\hat{W}_{i,j})=0$ and $\mathbb{E}_i(\hat{Q}(W_{i,j})^2)=\mathbb{E}_i(\hat{W}_{i,j}^2)$. Note that, due to regularization depends on the fixed parameters rather than dynamic activations, we can use weight re-parameterization to keep efficiency during inference.
\subsection{Perturbation Consistency Reconstruction}\label{sec:PCR}
Following QDrop~\cite{wei2022qdrop}, we employ block-wise reconstruction to determining quantization parameters by minimizing the feature distance before and after quantization. However, PD-quant~\cite{liu2023pd} points out that such an approach, which only considers local information, is prone to overfitting to the calibration samples. Moreover, randomly dropping the quantizated activations during PTQ and using fully quantizated activations during inference will cause unnegligible inconsistency between calibration and inference. 

We first introduce a global objective based on the analysis in~\cite{liu2023pd} to mitigate the overfitting:
\begin{equation}
\label{equ:l1}
\arg \min _{s} \mathcal{L}_{logits}(f(X^f_0), f_{l+1}(B_l^q(X^q_{l-1})))
\end{equation}
where $B_l^q$ represents the block being reconstructed, $f_{l+1}(\cdot)$ is the part that has not been quantified yet, $f(X^f_0)$ is FP output, and $X^f_0$ and $X^q_{l-1}$ denote input sample and the quantized feature from the previous reconstructed Block.

To further encourage the model to generate consistent predictions between the original and perturbed features, we propose perturbation consistency reconstruction loss. Specifically, we introduce a random perturbation $\epsilon_l$ to input feature $X^q_l$ of $l_{th}$ block. The classifier will output two prediction of the original feature $X^q_l$ and perturbed feature $\hat{X}^q_l=X^q_l+\epsilon_l$, denoted as $f_{l+1}(B_l^q(X^q_{l-1}))$ and $f_{l+1}(B_l^q(\hat{X}^q_{l-1}))$. We use Kullback-Leibler (KL) divergence as the objection to constrain the predicted probability of the perturbed features to be consistent with the original features, which can be formulated as:
\begin{equation}
\label{equ:l2}
\begin{aligned} 
\arg \min _{s} \mathcal{L}_{kl}(f_{l+1}(B_l^q(X^q_{l-1})), f_{l+1}(B_l^q(\hat{X}^q_{l-1})))
\end{aligned}
\end{equation}

Combining \equref{l1} and \equref{l2} and , we obtain perturbation consistency reconstruction loss:
\begin{equation}
\label{equ:pcrl}
\begin{aligned} 
\mathcal{L}_{pcr}=&\frac{1}{2}[\mathcal{L}_{logits}(f(X^f_0), f_{l+1}(B_l^q(X^q_{l-1})))\\
&+\mathcal{L}_{logits}(f(X^f_0), f_{l+1}(B_l^q(\hat{X}^q_{l-1})))]\\
&+\mathcal{L}_{kl}(f_{l+1}(B_l^q(X^q_{l-1})), f_{l+1}(B_l^q(\hat{X}^q_{l-1})))
\end{aligned}
\end{equation}

In actual implementation, excessively aggressive perturbations may result in significant differences between the predicted distributions of the original features and the perturbed features, thus causing the consistency constraint to fail. Therefore, careful selection of the perturbation strategy is necessary.

\subsection{Adaptive Balanced Loss}\label{sec:ABL}
In \secref{PEM} and \secref{PCR}, we propose PEM and PCR to enable the PTQ models to remain robust under minor perturbations in dynamic and ever-changing real-world scenarios. However, since the model needs to adapt online to sequentially arriving data, imbalanced class distributions can easily arise, resulting in catastrophic forgetting and unstable learning. Recent TTA research~\cite{Niu2022EfficientTM,marsden2024universal} is proposed to address the forgetting issue by employing sample reweighting for accuracy enhancement without considering class imbalance issue. 

To mitigate the impact of inter-class imbalance, we propose an adaptive balanced loss (ABL) to adjust the logits in loss during prediction. Unlike approaches~\cite{Zhao2023DELTADF,marsden2024universal}, ABL considers not only the frequency, but also the accumulated gradients of each class. Motivated by~\cite{Menon2020LongtailLV,He2024GradientRT}, we adjust the logits in ABL to achieve reweighting as follows:
\begin{equation}
\label{equ:abl}
\mathcal{L}_{\mathrm{ABL}}(y_i)=-\log \frac{e^{y_i+\log \pi_{y_i, t}}}{\sum_{j=1}^{\left|\mathcal{Y}^1\right|} e^{y_j+\log \pi_{y_j, t}}}
\end{equation}
where $\mathcal{Y}^1$ denotes class labels, and $\pi_{y,t}$ is the class prior $\mathbb{P}\left(y \mid \mathcal{S}_t\right)$ at time $t$. To tackle the dynamically changing and unknown input data stream, we use a momentum-updated class-wise vector, which is initiated with $\boldsymbol{\pi}=\left[\pi^1, \ldots \pi^{C}\right]$, to store the corresponding class-prior. To this end, we compute class-prior $\pi_{y,t}$ by:
\begin{equation}
\label{equ:pi}
\begin{aligned}
&\pi_{y_i, t}=\overbrace{\frac{\sum_{y\in \left|\mathcal{Y}^1\right|} \mathbf{1}\left(y_i=\mathcal{Y}^1\right)}{C}}^\textit{class frequency}\cdot \overbrace{\frac{\Phi_{i,t}}{\sum_{j=1}^{\left|\mathcal{Y}^1\right|} \Phi_{j,t}}}^\textit{class gradient}\\  
&where \quad \Phi_{i,t}=\sum_{n=1}^i\left\|\nabla_{\mathcal{L}_{c e}}\left(W_n\right)\right\|
\end{aligned}
\end{equation}
where $\mathbf{1}(\cdot)$ is the indicator function of label $y_i$ to count the frequency of each category, and $\Phi_{i,t}$ is the accumulated gradients of $i_{th}$ class, which is computed from the fully connected layer.

\section{Experiment}
In this section, we review the our TTAQ on several benchmark tasks to validate the effectiveness of our approach.

\subsection{Experimental Setup}

\textbf{Datasets.}
For the image classification task, we employ CIFAR10, CIFAR100, and ImageNet~\cite{imagenet15} as the source domain datasets, and CIFAR10-C, CIFAR100-C, and ImageNet-C/K~\cite{Hendrycks2019BenchmarkingNN} as the respective target domain datasets. These target domain datasets are designed to evaluate the robustness of classification networks, with each containing 15 types of corruption at 5 severity levels.
For the object detection task, we generate a corrupted version of COCO, labeled COCO-C, following the approach outlined in \cite{Michaelis2019BenchmarkingRI}, which introduces 15 types of corruptions. 

\textbf{Models and Implementation details.} All pretrained full precision models are obtained from \texttt{torchvision}. In the PTQ stage, we randomly sample 1024 images from the ImageNet training dataset as the calibration set. We maintain the same quantization settings and hyperparameters in our implementation as QDrop~\cite{wei2022qdrop}. 
In the CTTA stage, we follow~\cite{wang2020tent,Niu2022EfficientTM,marsden2024universal} to utilize a quantized W2A4 WideResNet~\cite{zagoruyko2016wide} and ResNeXt~\cite{xie2017aggregated} for CIFAR10-to-CIFAR10-C and CIFAR100-to-CIFAR100-C, respectively. For the ImageNet-to-ImageNet-C dataset, we employ quantized ResNet~\cite{he2016identity} with different bit widths for a comprehensive comparison. 

\begin{table*}[!h]
\renewcommand{\arraystretch}{1.2}
\centering
\small
\caption{Ablation study of PEM, PCR, and ABL is presented. All results for the continual test-time adaptation task are evaluated in ImageNet-C, with the largest corruption severity level 5.}
\label{tab:ablation}

\scalebox{0.74}{
\tabcolsep=4pt
\begin{tabular}{l|ccccccccccccccc|c}\toprule
\multicolumn{1}{l|}{Time} & \multicolumn{15}{l|}{$t\xrightarrow{\hspace*{14.5cm}}$}& \\ \hline
Method & \rotatebox[origin=c]{70}{Gaussian} & \rotatebox[origin=c]{70}{shot} & \rotatebox[origin=c]{70}{impulse} & \rotatebox[origin=c]{70}{defocus} & \rotatebox[origin=c]{70}{glass} & \rotatebox[origin=c]{70}{motion} & \rotatebox[origin=c]{70}{zoom} & \rotatebox[origin=c]{70}{snow} & \rotatebox[origin=c]{70}{frost} & \rotatebox[origin=c]{70}{fog}  & \rotatebox[origin=c]{70}{brightness} & \rotatebox[origin=c]{70}{contrast} & \rotatebox[origin=c]{70}{elastic} & \rotatebox[origin=c]{70}{pixelate} & \rotatebox[origin=c]{70}{jpeg} & Mean \\
\midrule
Baseline  &77.06 &66.96 &64.88 &75.14 &70.78 &63.66 &54.54 &57.18 &61.80 &47.88 &36.76 &70.54 &49.10 &46.06 &49.20 & 59.44 \\
+PEM      &74.28 &64.02 &63.50& \textbf{73.70}& 69.36& 62.22& 53.60& 55.66& 61.64& \textbf{46.50} & 36.12& 70.00& \textbf{47.78} & 44.76& \textbf{47.40} & 58.04 \\
+PEM+PCR  &73.60 &64.30 & 63.40 &73.88 &69.22 &\textbf{61.08} &53.22 &\textbf{55.40} &61.20 &46.78 &36.10 &69.62 &48.00 &\textbf{44.14} &47.64 & 57.88\\
+PEM+PCR+ABL  &\textbf{73.12} & \textbf{62.62} & \textbf{63.18} & 74.68 & \textbf{68.92} &61.26 & \textbf{53.08} &56.22 & \textbf{60.90} &46.56 & \textbf{35.56} & \textbf{68.60} & 47.92 &44.74 & 47.62& \underline{\textbf{57.67}} \\
\bottomrule
\end{tabular}
}
\end{table*}
\begin{table*}[t]
\caption{Classification error rate (\%) for the ImageNet-to-ImageNet-C and ImageNet-to-ImageNet-K online continual test-time
adaptation task on the highest corruption severity level 5.}
\small
\scalebox{0.82}{
\begin{tabular}{@{}l|cccccccccccc@{}}
\toprule
\multicolumn{1}{c}{\multirow{2.4}{*}{Method}} & \multicolumn{4}{c}{ResNet-50 on IN-C} & \multicolumn{4}{c}{ResNet-18 on IN-C} & \multicolumn{4}{c}{ResNet-50 on IN-K} \\ \cmidrule(lr{0pt}){2-5}  \cmidrule(lr{0pt}){6-9}  \cmidrule(lr{0pt}){10-13}
\multicolumn{1}{c}{} & W4A4 & \multicolumn{1}{c}{W3A3} & \multicolumn{1}{c}{W2A4} & \multicolumn{1}{c}{W2A2} & W4A4 & \multicolumn{1}{c}{W3A3} & \multicolumn{1}{c}{W2A4} & \multicolumn{1}{c}{W2A2} & W4A4 & \multicolumn{1}{c}{W3A3} & \multicolumn{1}{c}{W2A4} & \multicolumn{1}{c}{W2A2} \\ \midrule
Adaround & 59.08 & 66.88 & 69.48 & * & 66.34 & 70.46 & 72.57 & * & 67.02 & 73.30 & 76.65 & * \\
Brecq & 59.64 & 69.46 & 67.53 & 89.59 & 66.45 & 72.26 & 71.07 & 84.82 & 66.98 & 74.56 & 73.83 & 89.83 \\
Qdrop & 59.44 & 67.11 & 66.20  & 89.41 & 66.35  & 71.57 &70.46& 84.37 & 66.53 & 71.52 & 72.68 & 89.66 \\
TTAQ (ours) & \textbf{57.67} & \textbf{62.89}  & \textbf{62.05} & \textbf{80.59} & \textbf{65.17} &\textbf{69.49}&\textbf{68.99}& \textbf{82.71} & \textbf{66.24} & \textbf{69.91} &\textbf{ 69.52} & \textbf{83.80} \\ \bottomrule
\end{tabular}
}
\label{tab:imagenet}
\end{table*}
\subsection{Ablation Study}
To validate the efficiency and effectiveness of the proposed components, we report the classification error of W4A4 ResNet-50 on the ImageNet-to-ImageNet-C task with highest corruption severity level 5. As shown in \tabref{ablation}, we set the models quantized by QDrop and adapted by Tent as baseline and sequentially incorporate each of our contributions to better understand their individual effects on performance. Compared to the strong baseline, our TTAQ reduce the mean error by approximately 1.77\%. Specifically, when PEM is applied, the model to get 58.04\% mean error in W4A4 which is 1.4\% lower than the model quantized by qdrop. Further, the addition of the PCR reduced the model’s average error
rate from 58.04\% to 57.88\%, and when combined ABL, the average error further down to 57.67\%.

\begin{table*}[!h]
\renewcommand{\arraystretch}{1.2}
\centering
\small
\caption{Classification error rate~(\%) for the CIFAR100-to-CIFAR100C, CIFAR10-to-CIFAR10C online continual test-time adaptation task on the highest corruption severity level 5.}
\scalebox{0.74}{
\tabcolsep=4pt
\begin{tabular}{l|l|ccccccccccccccc|c}\toprule
& \multicolumn{1}{l|}{Time} & \multicolumn{15}{l|}{$t\xrightarrow{\hspace*{14.5cm}}$}& \\ \hline
\rotatebox[origin=c]{90}{Dataset}& Method & \rotatebox[origin=c]{70}{Gaussian} & \rotatebox[origin=c]{70}{Shot} & \rotatebox[origin=c]{70}{Impulse} & \rotatebox[origin=c]{70}{Defocus} & \rotatebox[origin=c]{70}{Glass} & \rotatebox[origin=c]{70}{motion} & \rotatebox[origin=c]{70}{Zoom} & \rotatebox[origin=c]{70}{Snow} & \rotatebox[origin=c]{70}{Frost} & \rotatebox[origin=c]{70}{Fog}  & \rotatebox[origin=c]{70}{Brightness} & \rotatebox[origin=c]{70}{Contrast} & \rotatebox[origin=c]{70}{Elastic} & \rotatebox[origin=c]{70}{Pixelate} & \rotatebox[origin=c]{70}{Jpeg} & Mean \\
\midrule
\multirow{5}{*}{\rotatebox[origin=c]{90}{Cifar100-C}}
& Full precision & 36.31 &31.89 &33.36 &25.13 &35.00 & 27.05 &24.22 &28.95 &28.85 &31.63 &23.14 &23.55 &30.81 &27.05 &34.10 &29.41
\\
\cline{2-18}
& Adaround  &39.30 &35.52 &38.07 &31.76 &38.93 &32.47 &29.43 &33.81 &33.49 &36.78 &27.20 &36.13 &35.40 &30.73 &36.56 & 34.35 \\
& Brecq      &40.26 &35.48 &38.23 &31.84 &38.87 &32.40 &29.56 &33.81 &33.44 &37.04 &27.37 &36.58 &35.62 &31.47 &37.49 & 34.73\\
& QDrop &40.44 &35.37 &37.68 &31.53 &39.16 &32.31 &29.32 &33.75 &33.10 &36.58 &27.19 &36.63 &35.58 &30.79 &37.07 & 34.53\\
& TTAQ (ours)   &\textbf{37.75}&\textbf{33.95}&\textbf{35.56} &\textbf{29.59} &\textbf{37.86} &\textbf{30.67} &\textbf{28.07} &\textbf{31.68} &\textbf{31.66} &\textbf{34.72} &\textbf{25.54} &\textbf{34.01} &\textbf{33.90} &\textbf{29.60} &\textbf{35.81} &\underline{\textbf{32.73}}\\
\midrule
\multirow{5}{*}{\rotatebox[origin=c]{90}{Cifar10-C}}
& Full precision & 23.62 &18.41&26.14&11.71&27.95&12.32&9.80&14.91&14.33&11.99&7.42&9.31&19.75&14.30&20.41&16.20
\\
\cline{2-18}
& Adaround  &24.34 &19.30 &26.25 &13.81 &28.59 &14.06 &12.12 &16.12 &15.46 &13.47 &8.40 &13.50 &20.91 &15.34 &20.25 & 17.46\\
& Brecq      &\textbf{24.31} &19.22 &26.12 & 14.23 &28.41 &14.16 &12.19 &15.76 &15.07 &13.33 &8.30 &13.63 &\textbf{20.23} &15.48 &19.77 & 17.43 \\
& QDrop &24.34 &18.99 &\textbf{26.04} &14.03 &\textbf{28.26} &13.88 &12.09 &15.67 &14.78 &13.30 &8.22 &13.48 &20.91 &15.65 &\textbf{19.67} & 17.31 \\
& TTAQ (ours)   &24.32&\textbf{18.63}& 27.05 &\textbf{13.28}& 29.27 &\textbf{13.37} &\textbf{11.08} &\textbf{14.80} &\textbf{14.72} &\textbf{13.24} &\textbf{8.09} &\textbf{12.00} & 20.42 &\textbf{14.69} & 20.48 & \underline{\textbf{17.08}}\\
\bottomrule
\end{tabular}
}
\label{tab:cifar}
\end{table*}
\subsection{Comparison on Classification Task}
\textbf{Results on ImagNet-C and ImagNet-K.} As shown in \tabref{imagenet}, we evaluate ResNet-18 and ResNet-50 models with different bit-widths and PTQ methods on various corrupted datasets. It is evident that the performance of traditional PTQ is highly sensitive to continuous domain variations. For instance, when the bit-width is W2A2, the average error of ResNet-50 on the ImageNet-C dataset increases by 29.95\% and 29.97\% for Brecq and QDrop respectively, while Adaround fails to predict in this setting altogether. In comparison, TTAQ achieves a reduction of 9\% and 8.82\% in average error compared to Brecq and QDrop, respectively. Furthermore, the issue of quantization parameter overfitting, as described in \secref{PCR}, is more pronounced in ResNet-50, which shows that the accuracy of W2A2 is even lower than that of R-18. However, our TTAQ method mitigates this problem and ensures a stable prediction across different models.

\begin{figure*}[t]
  \centering
  \vspace{0.14in}
  \includegraphics[width=1\linewidth]{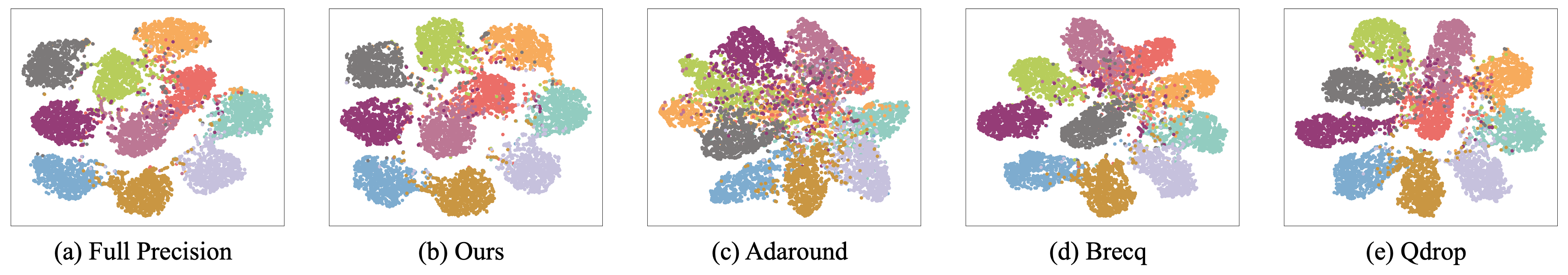}
  \vspace{-0.25in}
  \caption{T-SNE visualization of learned feature on quantized wider-resnet (W2A4), which reveal discriminative capability of classifier.}
  \label{fig:tsne}
\end{figure*}

\textbf{Results on Cifar100-C and Cifar10-C.} As presented in \tabref{cifar}, we evaluate the performance on CIFAR100-C and CIFAR10-C using quantized ResNext and WideResNet models with W2A4 bit-width.  We also provide a detailed the average error rate for each corruption type.
In CIFAR100-C, our TTAQ approach consistently outperforms the traditional PTQ method across all corruption types, resulting in an average error rate of 32.45\%, which is lower compared to the other three methods, with reductions of 1.78\%, 2.03\%, and 1.17\% respectively.
Similarly, in CIFAR10-C, our method demonstrates superior performance on most corruption types, resulting in an average error rate of 17.04\%. 
Additionally, as shown in \figref{tsne}, we provided t-SNE visualizations of the baselines and our method. The results demonstrated that the proposed TTAQ approach generated more discriminative features.
\begin{table*}[!t]
\renewcommand{\arraystretch}{1.2}
\centering
\small
\caption{Mean Average Precision~(\%) for the COCO-to-COCO-C online continual test-time adaptation task on the corruption severity level 1-5.}
\label{tab:continual-corruptions}
\scalebox{0.74}{
\tabcolsep=4pt
\begin{tabular}{l|l|ccccccccccccccc|c}\toprule
& \multicolumn{1}{l|}{Time} & \multicolumn{15}{l|}{$t\xrightarrow{\hspace*{14.5cm}}$}& \\ \hline
\rotatebox[origin=c]{90}{\#Bits}& Method & \rotatebox[origin=c]{70}{Gaussian} & \rotatebox[origin=c]{70}{Shot} & \rotatebox[origin=c]{70}{Impulse} & \rotatebox[origin=c]{70}{Defocus} & \rotatebox[origin=c]{70}{Glass} & \rotatebox[origin=c]{70}{motion} & \rotatebox[origin=c]{70}{Zoom} & \rotatebox[origin=c]{70}{Snow} & \rotatebox[origin=c]{70}{Frost} & \rotatebox[origin=c]{70}{Fog}  & \rotatebox[origin=c]{70}{Brightness} & \rotatebox[origin=c]{70}{Contrast} & \rotatebox[origin=c]{70}{Elastic} & \rotatebox[origin=c]{70}{Pixelate} & \rotatebox[origin=c]{70}{Jpeg} & mAP \\
\midrule
\rotatebox[origin=c]{90}{FP}
& -  &12.05 & 12.02 & 9.74 & 12.78 & 8.10 & 11.28 & 5.59 & 12.64 & 15.56 & 20.68 & 21.82 & 15.36 & 15.14 & 9.54 & 10.68 & 12.86\\
\midrule
\multirow{4}{*}{\rotatebox[origin=c]{90}{W4A4}}& Adaround  &4.63 & 4.57 & 3.02 & 3.91 & 2.73 & 4.45 & 2.07 & 4.47 & 5.47 & 6.47 & 11.00 & 3.77 & 7.86 & 3.47 & 4.85 & 8.85 \\
& Brecq      &10.48 &10.51 &8.50 &10.44 &7.63 &9.89 &4.69 &11.64 &\textbf{13.94} &16.04 &20.62 &13.94 &14.57 &9.33 &\textbf{11.67} & 11.32 \\
& QDrop &10.85 & 10.94 & 9.03 & 10.47 & 7.65 & 10.14 & 4.81 & 11.43 & 13.93 & 16.84 & 20.70 & 10.74 & 14.38 & 9.59 & 11.53 & 11.54\\
& TTAQ (ours)  &\textbf{10.94} & \textbf{11.15} & \textbf{9.14} & \textbf{11.23} & \textbf{7.67} & \textbf{10.44} &\textbf{ 4.99} & \textbf{11.97} & \textbf{14.39} & \textbf{17.63} & \textbf{21.21} & 11.59 & \textbf{14.71} & \textbf{9.72} & 11.24 & \underline{\textbf{11.87}}\\
\midrule
\multirow{4}{*}{\rotatebox[origin=c]{90}{W2A4}}& Adaround  &4.46 & 4.60 & 3.26 & 4.87 & 3.85 & 4.80 & 2.24 & 5.25 & 7.12 & 6.47 & 11.44 & 3.58 & 7.92 & 4.63 & 5.49 & 5.33 \\
& Brecq      &6.80 & 7.00 & 5.34 & 6.97 & 5.82 & 6.94 & 3.07 & 7.70 & 9.89 & 10.74 & 16.28 & 5.88 &12.29 & 7.28 &9.21 & 8.08 \\
& QDrop & 7.30 & 7.36 & 5.60 & 7.41 & 5.97 & 7.45 & 3.23 & 8.04 & 10.09 & 11.93 & 16.78 & 6.97 & 12.13 & 7.12 & 8.70 & 8.41\\
& TTAQ (ours)  &\textbf{9.31} & \textbf{9.33} & \textbf{7.46} &\textbf{ 9.90} & \textbf{6.38} & \textbf{8.86} & \textbf{3.95} & \textbf{10.21} & \textbf{12.69} & \textbf{15.23} & \textbf{19.00} & \textbf{9.62} & \textbf{12.92} &\textbf{ 8.55} & \textbf{9.99} & \underline{\textbf{10.23}}\\
\bottomrule
\end{tabular}
}
\label{tab:coco}
\end{table*}
\subsection{Comparison on Object Detection Task}
The experimental results of the object detection task, which conducted on the COCO-C dataset with the Faster RCNN framework, are reported in \tabref{coco}. We set the quantized ResNet-50 with W4A4 and W2A4 bit-width as the backbone, while keeping other components in full precision.
In the W4A4 bit-width setting, our method outperforms Brecq and QDrop by 0.55\% and 0.33\% mAP, respectively. Furthermore, in the W2A4 bit-width setting, our method demonstrates an improvement in performance over Brecq and QDrop by 2.15\% and 1.82\% mAP, respectively. Similar to the results in the classification task, our method maintains a more stable performance even at lower bit-widths.

\section{Conclusion}
In this paper, we introduce TTAQ, a novel approach for stable post-training quantization in continuous domain adaptation. TTAQ incorporates Perturbation Error Mitigation, which analyzes error propagation based on signal-to-noise ratio and implements a weight regularization scheme to minimize the impact of input perturbations. Additionally, TTAQ introduces Perturbation Consistent Reconstruction to ensure consistent predictions when there is a small perturbation in calibration. Furthermore, Adaptive Balanced Loss is proposed to address class imbalance by adjusting sample weights according to class-wise factors, thereby enhancing the learning of unbiased classifiers. Exhaustive experiments across various vision tasks validate the superiority of TTAQ, demonstrating significant performance improvements over existing PTQ methods in continuous domain adaptation. 

\clearpage
\bibliographystyle{plainnat}
\bibliography{main}

\end{document}